\documentclass[runningheads]{llncs}

\usepackage{tikz}
\usepackage{graphicx}
\usepackage{algorithm}
\usepackage[noend]{algpseudocode}
\begin{document}
\title{Shapley Chains: Extending Shapley Values to Classifier Chains
}
\author{Célia Wafa Ayad\inst{1,2}\and
Thomas Bonnier\inst{2}\and
Benjamin Bosch\inst{2}\and
Jesse Read\inst{1}
}
\authorrunning{C. AYAD et al.}
\institute{LIX, \'Ecole Polytechnique, Institut Polytechnique de Paris \and
Soci\'et\'e G\'en\'erale\\
}
\maketitle              
\begin{abstract}
  In spite of increased attention on explainable machine learning models, explaining multi-output predictions has not yet been extensively addressed. 
  Methods that use Shapley values to attribute feature contributions to the decision making are one of the most popular approaches to explain local individual and global predictions.
  By considering each output separately in multi-output tasks, these methods fail to provide complete feature explanations.
  We propose Shapley Chains to overcome this issue by including label interdependencies in the explanation design process. 
  Shapley Chains assign Shapley values as feature importance scores in multi-output classification using  classifier chains, by separating the direct and indirect influence of these feature scores.
  Compared to existing methods, this approach allows to attribute a more complete feature contribution to the predictions of multi-output classification tasks. 
  We provide a mechanism to distribute the hidden contributions of the outputs with respect to a given chaining order of these outputs. 
  Moreover, we show how our approach can reveal indirect feature contributions missed by existing approaches. 
  Shapley Chains help to emphasize the real learning factors in multi-output applications and allows a better understanding of the flow of information through output interdependencies in synthetic and real-world datasets.
  
  \keywords{Machine Learning Explainability  \and Classifier Chains \and Multi-Output Classification \and Shapley Values.}
\end{abstract}

\section{Introduction}

A multi-output model predicts several outputs from one input. This is an important learning problem for decision-making involving multiple factors and complex criteria in the real-world scenarios, such as in healthcare, the prediction of multiple diseases for individual patients. 
Classifier chains \cite{readClassifierChainsReview2021} is one such approach for multi-output classification, taking output dependencies into account by connecting individual base classifiers, one for each output.
The order of output nodes and the choice of the base classifiers are two parameters yielding different predictions thus different explanations for the given classifier chain.

To address the lack of transparency in existing machine learning models, solutions such as SHAP \cite{lundbergUnifiedApproachInterpreting2017a}, LIME \cite{ribeiroWhyShouldTrust2016a}, DEEPLIFT \cite{shrikumarLearningImportantFeatures2019a} and Integrated Gradients \cite{sundararajanAxiomaticAttributionDeep2017a} have been proposed. 
Using Shapley values \cite{rozemberczkiShapleyValueClassifiers2021a} is one approach to attribute feature importance in machine learning. The framework SHAP \cite{lundbergUnifiedApproachInterpreting2017a} provides Shapely values used to explain model predictions, by computing feature marginal contributions to all subsets of features.
This theoretically well founded approach provides instance-level explanations and a global interpretation of model predictions by combining these local (instance-level) explanations.

However, these methods are not suitable for multi-output configurations, especially when these outputs are interdependent. In addition, the SHAP framework provides separate feature importance scores only for independent multi-output classifiers. By assuming the independence of outputs, one ignores the indirect connections between features and outputs, which leads to assigning incomplete feature contributions, thus an inaccurate explanation of the predictions.

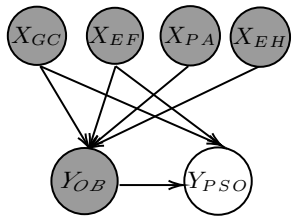
\begin{figure}[h!]
\begin{center}
\tikzset{every picture/.style={line width=0.75pt}} 
\begin{tikzpicture}[x=0.5pt,y=0.5pt,yscale=-1,xscale=1]
\draw    (526,61) -- (452.54,121.73) ;
\draw [shift={(451,123)}, rotate = 320.42] [color={rgb, 255:red, 0; green, 0; blue, 0 }  ][line width=0.75]    (10.93,-3.29) .. controls (6.95,-1.4) and (3.31,-0.3) .. (0,0) .. controls (3.31,0.3) and (6.95,1.4) .. (10.93,3.29)   ;
\draw    (414,61) -- (449.98,121.28) ;
\draw [shift={(451,123)}, rotate = 239.17] [color={rgb, 255:red, 0; green, 0; blue, 0 }  ][line width=0.75]    (10.93,-3.29) .. controls (6.95,-1.4) and (3.31,-0.3) .. (0,0) .. controls (3.31,0.3) and (6.95,1.4) .. (10.93,3.29)   ;
\draw    (471,61) -- (451.61,121.1) ;
\draw [shift={(451,123)}, rotate = 287.88] [color={rgb, 255:red, 0; green, 0; blue, 0 }  ][line width=0.75]    (10.93,-3.29) .. controls (6.95,-1.4) and (3.31,-0.3) .. (0,0) .. controls (3.31,0.3) and (6.95,1.4) .. (10.93,3.29)   ;
\draw    (580,61) -- (452.8,122.13) ;
\draw [shift={(451,123)}, rotate = 334.33] [color={rgb, 255:red, 0; green, 0; blue, 0 }  ][line width=0.75]    (10.93,-3.29) .. controls (6.95,-1.4) and (3.31,-0.3) .. (0,0) .. controls (3.31,0.3) and (6.95,1.4) .. (10.93,3.29)   ;
\draw    (471,61) -- (549.4,119.8) ;
\draw [shift={(551,121)}, rotate = 216.87] [color={rgb, 255:red, 0; green, 0; blue, 0 }  ][line width=0.75]    (10.93,-3.29) .. controls (6.95,-1.4) and (3.31,-0.3) .. (0,0) .. controls (3.31,0.3) and (6.95,1.4) .. (10.93,3.29)   ;
\draw    (414,61) -- (549.17,120.2) ;
\draw [shift={(551,121)}, rotate = 203.65] [color={rgb, 255:red, 0; green, 0; blue, 0 }  ][line width=0.75]    (10.93,-3.29) .. controls (6.95,-1.4) and (3.31,-0.3) .. (0,0) .. controls (3.31,0.3) and (6.95,1.4) .. (10.93,3.29)   ;
\draw    (474,151) -- (520,151) ;
\draw [shift={(522,151)}, rotate = 180] [color={rgb, 255:red, 0; green, 0; blue, 0 }  ][line width=0.75]    (10.93,-3.29) .. controls (6.95,-1.4) and (3.31,-0.3) .. (0,0) .. controls (3.31,0.3) and (6.95,1.4) .. (10.93,3.29)   ;
\draw  [fill={rgb, 255:red, 155; green, 155; blue, 155 }  ,fill opacity=1 ]  (527.5, 38) circle [x radius= 22.47, y radius= 22.47]   ;
\draw (527.5,38) node    {$X_{P}{}_{A}$};
\draw  [fill={rgb, 255:red, 155; green, 155; blue, 155 }  ,fill opacity=1 ] (447.5, 148) circle [x radius= 25.06, y radius= 25.06]   ;
\draw (447.5,148) node  [xslant=0.32]  {$Y_{O}{}_{B}$};
\draw    (548.5, 148) circle [x radius= 25.5, y radius= 25.5]   ;
\draw (548.5,148) node    {$Y_{P}{}_{S}{}_{O}$};
\draw  [fill={rgb, 255:red, 155; green, 155; blue, 155 }  ,fill opacity=1 ]  (470.5, 38) circle [x radius= 21.22, y radius= 21.22]   ;
\draw (470.5,38) node    {$X_{E}{}_{F}$};
\draw  [fill={rgb, 255:red, 155; green, 155; blue, 155 }  ,fill opacity=1 ]  (580.5, 39) circle [x radius= 22.47, y radius= 22.47]   ;
\draw (580.5,39) node    {$X_{E}{}_{H}$};
\draw  [fill={rgb, 255:red, 155; green, 155; blue, 155 }  ,fill opacity=1 ]  (411.5, 38) circle [x radius= 22.05, y radius= 22.05]   ;
\draw (411.5,38) node    {$X_{GC}$};
\end{tikzpicture}
\caption{
An example of a multi-output task: predicting $Y$-outputs from $X$-features.
A classifier chain uses the first output $Y_{\mathsf{OB}}$ as an additional feature to predict 
the second output $Y_{\mathsf{PSO}}$. } \label{fig:health}
\end{center}
\end{figure}

Fig.~\ref{fig:health} is a graphical representation of a classifier chain: patients with two conditions, 
obesity ($Y_{\mathsf{OB}}$) and psoriasis ($Y_{\mathsf{PSO}}$), given four features: genetic components ($X_{\mathsf{GC}}$), environmental factors ($X_{\mathsf{EF}}$), physical activity ($X_{\mathsf{PA}}$) and eating habits ($X_{\mathsf{EH}}$). 
From a clinical point of view, all factors $X$ are associated with both conditions $Y$, obesity and psoriasis. However, since obesity is a strong feature for predicting psoriasis \cite{jensenPsoriasisObesity2016} (indeed, a motivating factor for using such a model is that predictive accuracy can be improved by incorporating outputs as features), it may mask the effects of other features. Namely, $X_\mathsf{PA}$ and $X_\mathsf{EH}$ will be found by methods as SHAP applied to each output separately to have zero contribution towards predicting $Y_\mathsf{PSO}$, and one might interpret that psoriasis is mainly affected by factors which cannot be modified by the patient (environment and genetics). The \emph{indirect} effects (physical activity and eating habits) will not be detected or explained.

We propose Shapley Chains to address this limitation of incomplete attribution of feature importance in multi-output classification tasks by taking into account the relationships between outputs and distributing their importance among the features with respect to a given order of these outputs.
Calculating the Shapley values of outputs helps to better understand the importance of the chaining that connects these outputs and to visualize this relationship impact on the prediction of subsequent outputs in the chain. For these subsequent outputs, the computation of the Shapley values of the associated outputs shows the indirect influence of some features through the chain, which is generally not intuitive and missed by existing work.
Our method will successfully explain these \emph{indirect} effects. 
By attributing importance to the features $X_{\mathsf{PA}}$ and $X_{\mathsf{EH}}$, Shapley Chains will help doctors to emphasize the importance of eating healthy and practicing physical activities in order to prevent and better cure psoriasis instead of blaming only genetics and exterior environmental factors.

This paper addresses the problem of attributing feature contributions in multi-output classification tasks with classifier chains when outputs are interdependent. Our contribution in this paper is resumed to :
\begin{itemize}
    \item We propose Shapley Chains, a novel post-hoc model agnostic explainability method designed for multi-output classification task using classifier chains. 
    \item Shapley Chains attribute feature importance to all features that directly or indirectly contribute to the prediction of a given output, by tracking all the related outputs in the given chain order.
    \item Compared to existing methods, we show a more complete distribution of feature importance scores in multi-output synthetic and real-world datasets.
\end{itemize}

We devote Section 2 to a background and related work. In Section 3, we detail our proposed method Shapley Chains.
Finally in Section 4, we run experiments on a synthetic and real-world datasets. 
The results of our method compared to SHAP values applied to independent classifiers are then discussed.

\section{Background and Related Work}
In this section we review multi-output classification, output dependencies, classifier chains and Shapley values to serve as a background for the rest of this paper. The notation we used is summarized in the next table.

\begin{table}[h!]
\centering
\caption{Notation} \label{tab:not}
\begin{tabular}{|l|l|}
\hline
 \textbf{Notation} &  \textbf{Meaning}\\
\hline
\textbf{x} & a given instance vector\\
\textbf{y} & a given output vector\\
 $x_i$ & the $i^{th}$ feature of instance \textbf{x}\\
 $y_j$ & the $j^{th}$ output\\
 $X$ & the feature space of $x_i$\\
 $Y$ & the output space of $y_j$\\
 $n$ & the number of features for each instance \textbf{x}\\
 $m$ & the number of outputs\\
\hline
\end{tabular}
\end{table}

\label{sec:dependence}
\subsection{Multi-output classification and output dependencies}

A multi-output classifier $\mathsf{H}$ is a mapping function that for a given instance \textbf{x}=$ \{x_1, x_2, ...,x_{n}\}$, such that $\textbf{x} \in X$, it learns a vector of base classifiers $\mathsf{H}($\textbf{x}$)= h_1($\textbf{x}$),h_2($\textbf{x}$),...,h_{m}($\textbf{x}$)$ and returns a vector of predicted values $\textbf{y}=\{y_1, y_2, ..., y_{m}\}$
, with $y_j \in \{0,1\}$ and $\textbf{y} \in Y$. 

In real-world applications, outputs can be dependent or independent. Designing classifiers that incorporate these output dependencies makes it possible to better represent the relationships in the data (between outputs, therefore between features and outputs). There are two types of output dependencies wrt subsequent outputs; namely marginal independencies,  
$P(\textbf{y})= \prod_{j=1}^{m}{P(y_j)}$, and 
conditional output dependencies: 
\begin{equation}\label{pcc}
    P(\textbf{y}|\textbf{x})= \prod_{j=1}^{m}{P(y_j|X, y_1,..., y_{j-1})}
\end{equation} 

In this article, we focus on output conditional dependencies.
The nature of the relationship between features and outputs and between outputs is not restricted to causality. Therefore, no prior knowledge of the causal graph is necessary. This specific subject is partially covered in Shapley Flow \cite{wangShapleyFlowGraphbased2021}, which is designed for single-output tasks.

\subsection{Classifier chains}

A classifier chain is one multi-output method that learns $m$ classifiers (one classifier for each output, also referred as base classifier). 
All the classifiers are linked in a chain. The chaining method passes output information between classifiers, allowing this method to take into account output dependencies \cite{readClassifierChainsMultilabel2011} when learning a given output in the chaining. 

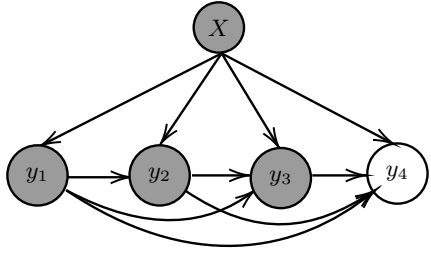
\begin{figure}
\begin{center}
\tikzset{every picture/.style={line width=0.9pt}} 
\begin{tikzpicture}[x=0.7pt,y=0.7pt,yscale=-1,xscale=1]
\draw    (684,55) -- (587.78,104.09) ;
\draw [shift={(586,105)}, rotate = 332.97] [color={rgb, 255:red, 0; green, 0; blue, 0 }  ][line width=0.75]    (10.93,-3.29) .. controls (6.95,-1.4) and (3.31,-0.3) .. (0,0) .. controls (3.31,0.3) and (6.95,1.4) .. (10.93,3.29)   ;
\draw    (684,55) -- (652.12,102.34) ;
\draw [shift={(651,104)}, rotate = 303.96] [color={rgb, 255:red, 0; green, 0; blue, 0 }  ][line width=0.75]    (10.93,-3.29) .. controls (6.95,-1.4) and (3.31,-0.3) .. (0,0) .. controls (3.31,0.3) and (6.95,1.4) .. (10.93,3.29)   ;
\draw    (684,55) -- (775.23,103.07) ;
\draw [shift={(777,104)}, rotate = 207.78] [color={rgb, 255:red, 0; green, 0; blue, 0 }  ][line width=0.75]    (10.93,-3.29) .. controls (6.95,-1.4) and (3.31,-0.3) .. (0,0) .. controls (3.31,0.3) and (6.95,1.4) .. (10.93,3.29)   ;
\draw    (684,55) -- (713.96,104.29) ;
\draw [shift={(715,106)}, rotate = 238.71] [color={rgb, 255:red, 0; green, 0; blue, 0 }  ][line width=0.75]    (10.93,-3.29) .. controls (6.95,-1.4) and (3.31,-0.3) .. (0,0) .. controls (3.31,0.3) and (6.95,1.4) .. (10.93,3.29)   ;
\draw    (601,122) -- (633,122) ;
\draw [shift={(635,122)}, rotate = 180] [color={rgb, 255:red, 0; green, 0; blue, 0 }  ][line width=0.75]    (10.93,-3.29) .. controls (6.95,-1.4) and (3.31,-0.3) .. (0,0) .. controls (3.31,0.3) and (6.95,1.4) .. (10.93,3.29)   ;
\draw    (668,121) -- (697,121) ;
\draw [shift={(699,121)}, rotate = 180] [color={rgb, 255:red, 0; green, 0; blue, 0 }  ][line width=0.75]    (10.93,-3.29) .. controls (6.95,-1.4) and (3.31,-0.3) .. (0,0) .. controls (3.31,0.3) and (6.95,1.4) .. (10.93,3.29)   ;
\draw    (733,121) -- (760,121) ;
\draw [shift={(762,121)}, rotate = 180] [color={rgb, 255:red, 0; green, 0; blue, 0 }  ][line width=0.75]    (10.93,-3.29) .. controls (6.95,-1.4) and (3.31,-0.3) .. (0,0) .. controls (3.31,0.3) and (6.95,1.4) .. (10.93,3.29)   ;
\draw    (600,129) .. controls (644.1,152.52) and (676.68,148.19) .. (700.55,130.12) ;
\draw [shift={(702,129)}, rotate = 501.63] [color={rgb, 255:red, 0; green, 0; blue, 0 }  ][line width=0.75]    (10.93,-3.29) .. controls (6.95,-1.4) and (3.31,-0.3) .. (0,0) .. controls (3.31,0.3) and (6.95,1.4) .. (10.93,3.29)   ;
\draw    (664,129) .. controls (703.4,155.6) and (733.1,151.14) .. (765.52,129.98) ;
\draw [shift={(767,129)}, rotate = 506.31] [color={rgb, 255:red, 0; green, 0; blue, 0 }  ][line width=0.75]    (10.93,-3.29) .. controls (6.95,-1.4) and (3.31,-0.3) .. (0,0) .. controls (3.31,0.3) and (6.95,1.4) .. (10.93,3.29)   ;
\draw    (600,129) .. controls (666.33,173.55) and (723.84,164.19) .. (765.74,130.04) ;
\draw [shift={(767,129)}, rotate = 500.19] [color={rgb, 255:red, 0; green, 0; blue, 0 }  ][line width=0.75]    (10.93,-3.29) .. controls (6.95,-1.4) and (3.31,-0.3) .. (0,0) .. controls (3.31,0.3) and (6.95,1.4) .. (10.93,3.29)   ;
\draw   [fill={rgb, 255:red, 155; green, 155; blue, 155 }  ,fill opacity=1 ]  (682.5, 41) circle [x radius= 13.6, y radius= 13.6]   ;
\draw (682.5,41) node    {$X$};
\draw    [fill={rgb, 255:red, 155; green, 155; blue, 155 }  ,fill opacity=1 ] (584.5, 121) circle [x radius= 16.28, y radius= 16.28]   ;
\draw (584.5,121) node    {$y_{1}$};
\draw    [fill={rgb, 255:red, 155; green, 155; blue, 155 }  ,fill opacity=1 ] (650.5, 121) circle [x radius= 16.28, y radius= 16.28]   ;
\draw (650.5,121) node    {$y_{2}$};
\draw   [fill={rgb, 255:red, 155; green, 155; blue, 155 }  ,fill opacity=1 ]  (716, 122.44) circle [x radius= 16.28, y radius= 16.28]   ;
\draw (716,122.44) node    {$y_{3}$};
\draw    (779, 120) circle [x radius= 16.28, y radius= 16.28]   ;
\draw (779,120) node    {$y_{4}$};
\end{tikzpicture}
\caption{One example of a classifier chain structure } \label{fig:CC}
\end{center}
\end{figure}

This method is exactly an expression of Eq.~\ref{pcc}, if expressed according to the chain rule of probability (i.e., Fig.~\ref{fig:CC} as a probabilistic graphical model representation). That is one reason why conditional dependencies are interesting in this context. However, a classifier chain is not faithful to a `proper' inference procedure, and rather takes a greedy approach to inference, plugging in predictions as observations; and proceeds much as a forward pass across a neural network. This creates some ambiguity between how much effect is gained from probabilistic dependence (as a probabilistic graphical model would) and feature effect (as one encounters via the latent layers of deep learning). Although discussion has been ongoing e.g., \cite{readClassifierChainsReview2021,readClassifierChainsMultilabel2011}, there is not yet a consistent understanding in practice of what role a prediction plays as a feature to another label. By propagating output contributions among the features, Shapley Chains help to clarify these prediction roles, and confirm which outputs are interdependent using the Shapley value described in the next section. 

\subsection{Shapley values} \label{sv}
The Shapley value expresses the contribution of feature $x_i$, to predict output $y_j$ as a weighted sum: 
\begin{equation}\label{first_eq}
     \phi_{y_{j}} x_{i} \ =\ \sum\limits _{S\subseteq X\backslash \{i\} \ }\frac{|S|!\ ( |X|\ -\ |S|\ -1) !}{|X|!} \ [f_x\ ( S\cup \{i\}) \ -\ f_x\ (S)]
\end{equation}

Where $S \subseteq X$, 
and $f_x$ is the value function that defines each feature's contribution to each subset $S$. It computes each feature's average added value to each combination of features when making a prediction for instance \textbf{x}.

Additivity is one axiom of a fair attribution mechanism that is satisfied by the Shapley value. It finds a good interpretation in multi-output classification. Consider two prediction tasks ($X$, $f$), ($X$, $g$) composed of the same set of features. We create a coalition prediction task $(X, f + g)$ by adding the two previous prediction tasks in the following way: $(f + g)(S)$ = $f(S) + g(S)$ for all $S \subseteq X$. The additivity axiom states that the allocation of the prediction $(X, f + g)$ will be equal to the sum of the allocations of the two original prediction tasks. One should note that in this definition, we assume that the two prediction tasks are completely independent meaning that feature contributions to one prediction has no effect on the second one, which is not always the case because in real-world applications tasks are more often interdependent. 
One approach we propose is to use classifier chains 
because it permits to represent these relationships by introducing different chaining orders of these outputs. The overall feature Shapley values for a classifier chain can be calculated by marginalizing over all possible output chain structures. $\forall c \in \mathsf{C}$, the Shapley value of $x_{i}$ in Eq.~\ref{first_eq} can be written as follows:
\begin{equation}\label{other_eq}
     \phi_{y_{j} x_{i} } = \frac{1}{|\mathsf{C}|} \sum\limits _{ c\subseteq \mathsf{C} }  \phi_{y_{j}^c} x_{i} 
\end{equation}
with $\phi_{y_{j}^c}$ being the contribution of feature $x_i$ to the prediction of $y_j$ with respect to the given chaining order $c$. For the matter of simplicity, we use $\phi_{y_{j}}$ to refer to $\phi_{y_{j}^c}$ in the rest of this paper.
We report feature contribution for each chain structure independently to show the impact of different chaining orders and the marginalization over these orders in Section \ref{impact}.

\subsection{Related work}

The explainability of machine learning is an active research topic in the recent years. Several contributions have been made to explain single-output models and predictions. 
Inspecting feature importance scores of existing models is an intuitive approach that has served for many studies. These feature importance scores are either derived directly from feature weights in a linear regression for instance, or learned from feature permutations based on the decrease in model performance. 
Other more complex methods like LIME \cite{ribeiroWhyShouldTrust2016a} learn a surrogate model locally (around a given instance) in order to explain the predictions of the initial model with simple and interpretable models like decision trees. On the other hand, DeepLift \cite{shrikumarLearningImportantFeatures2019a}, Integrated gradient \cite{sundararajanAxiomaticAttributionDeep2017a} and LRP \cite{montavonLayerWiseRelevancePropagation2019} are some neural network specific methods proposed to explain deep neural networks. 

The SHAP framework is one popular method attributing Shapley values as feature contributions. It provides a wide range of model-specific and model-agnostic explainers. 
Researchers have also proposed other Shapley value inspired methods incorporating feature interactions in the explanation process. For example, asymmetric Shapley values \cite{fryeAsymmetricShapleyValues2021} incorporates causal knowledge into model explanations. This method attributes importance scores to features that do not directly participate in the prediction process (confounders), but fails to capture all direct feature contribution. On the other hand, on manifold Shapley values \cite{fryeShapleyExplainabilityData2021} focus on better representing the out of coalition feature values but provides misleading interpretation of feature contributions. 
Wang et al. \cite{wangShapleyFlowGraphbased2021} have proposed Shapley Flow, providing both direct and indirect feature contributions when a causal graph is provided. Resuming feature interactions to causality and assuming the causal graph is provided and accurate are two downsides of this method. 
These methods significantly contributed to advancing the explainability of machine learning models but none of them have tackled multi-output problems, more specifically when outputs are interdependent. Shapley Chains address this limitation. 

\section{Proposed Method: Shapley Chains}
In this section, we introduce our approach to compute direct and indirect feature Shapley values for a classifier chain model. Note that our proposed method is model-agnostic, meaning that our computations do not depend directly on the chosen base learner used by the classifier chain. 

\begin{figure}
\centering
\includegraphics[scale=0.3]{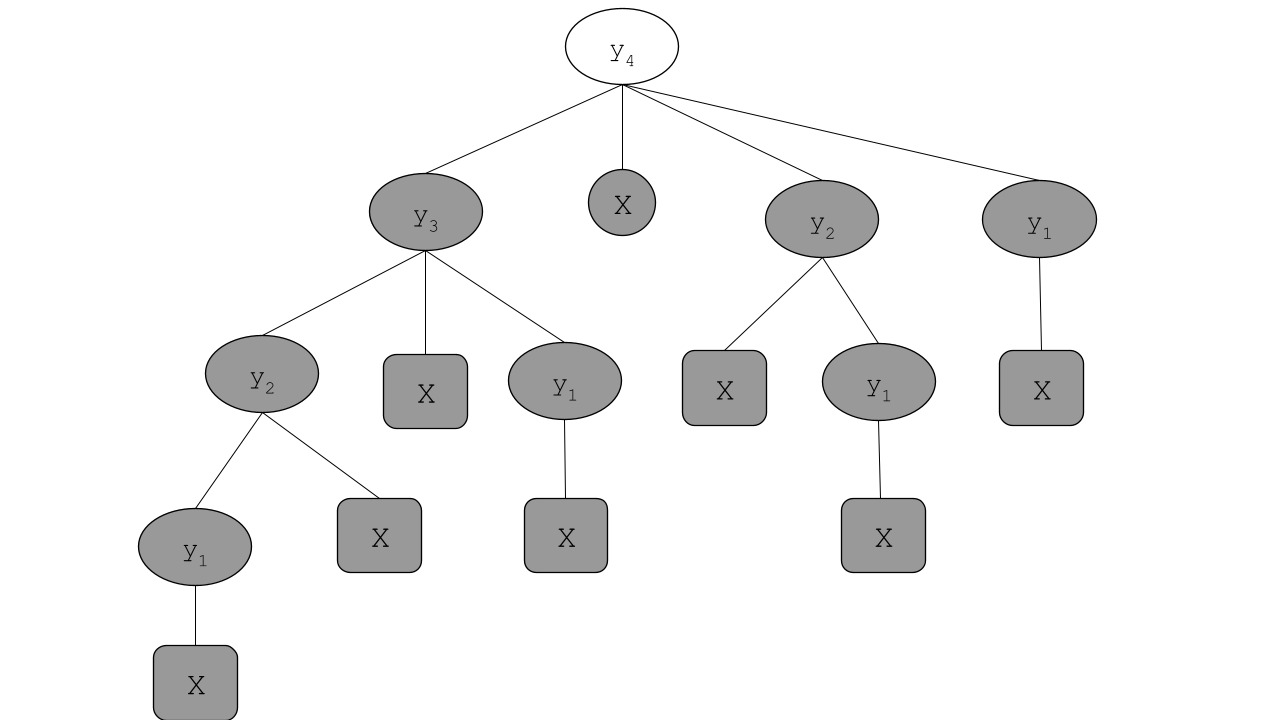}
\caption{Representation of direct and indirect contributions for a dataset with 4 outputs ($y_1$, $y_2$, $y_3$ and $y_4$). For example: the 4th output $y_4$ has 7 indirect Shapley values (7 paths ending with square leave) and one direct Shapley value (one path ending with a circle leaf).}
\label{fig:indirect}
\end{figure}

We want to compute feature contributions to the prediction of each output $ y_j\in Y$ for each instance \textbf{x}. For example, Fig.~\ref{fig:indirect} shows the direct and indirect contributions of $x_i$ to predict output $y_4$ given in Fig.~\ref{fig:CC}. 
In the next two sections, we detail the computations of the Shapley value of each feature to predict each output. We refer to these Shapley values as direct and indirect feature contributions.

\subsubsection{Direct contributions}

The direct contributions are computed for features and outputs as in Eq.~\ref{first_eq}.
Consider again the example of patients with the two conditions: psoriasis and obesity. For both $Y_{\mathsf{OB}}$ and $Y_{\mathsf{PSO}}$, we use the framework SHAP in order to compute the Shapley value of each feature : $X_{\mathsf{GC}}$, $X_{\mathsf{EF}}$, $X_{\mathsf{PA}}$ and $X_{\mathsf{EH}}$.
This will attribute non zero Shapley values to $X_{\mathsf{GC}}$ and $X_{\mathsf{EF}}$ to predict $Y_{\mathsf{OB}}$ and $Y_{\mathsf{PSO}}$ separately. On the other hand, $X_{\mathsf{EF}}$ and $X_{\mathsf{PA}}$ will have non-zero Shapley values to predict $Y_{\mathsf{OB}}$ and zero values for the prediction of $Y_{\mathsf{PSO}}$. 
The classifier chain method will add $Y_{\mathsf{OB}}$ to the feature set to predict $Y_{\mathsf{PSO}}$. By running the SHAP framework on this new set, $Y_{\mathsf{OB}}$ will have a non zero Shapley value because it is dependent to $Y_{\mathsf{PSO}}$. This Shapley value will be attributed to the features that are correlated to $Y_{\mathsf{OB}}$. 
The attribution mechanism of direct feature (and output) contributions can be generalized to the classifier $\mathsf{H}$ with $m$ base classifiers as shown in Algorithm~\ref{alg1}.

\begin{algorithm}
\caption{Computing direct feature contributions}\label{alg1}
\begin{algorithmic}[1]
\Procedure{diContribution}{$X, Y, H$}   \Comment{features, outputs, classifier chain model}
    \State $i=j=0$
    \State $\Phi$=[]
    \While{$j<len(Y)$} 
        \While{$i<len(X)$}
            \State $\Phi_{y_{j}} x_{i}  \leftarrow SHAP(X, y_j, H)$  \Comment{Shapley values of inputs wrt each output}
            \State append $y_j$ to $X$
            \State append $\Phi_{y_{j}} x_{i}$ to $\Phi$
        \EndWhile 
    \EndWhile  
    \State return $\Phi$ \Comment{$\Phi$ contains features and outputs Shapley values}
\EndProcedure

\end{algorithmic}
\end{algorithm}

For the first output $y_1$, we calculate the Shapley value of each feature according to Eq.~\ref{first_eq}, as done in the SHAP framework. 
This marginal value of all possible subsets to which the feature can be associated to is the feature's contribution to predict the first output $y_1$.
For the second output $y_2$, we append the predictions $y_1$ made by the first classifier $h_1$ to the features set, and we train a second classifier $h_2$ to learn the second output $y_2$. We again use the SHAP framework to assign Shapley values to features and the first output $y_1$. Here, the feature set includes the first prediction. 
We perform the same steps for each remaining output. At each step, we calculate the Shapley values for features and previous predicted outputs that are linked via the chaining to the current output. At the final step, the feature set will contain $n$ features and $m$ outputs: $X=\{x_1, x_2, ..., x_n, y_1, y_2,..., y_{m}\}$.

\subsubsection{Indirect contributions}

The indirect contribution $\Phi_{indirect} y_j(x_i)$ of $x_i$ to predict $y_j$ is the weighted sum of the direct contributions of all $y_k \in Y$ that are chained to $y_j$. $\Phi_{indirect} y_j(x_i)$ is computed according to the Eq.~\ref{four_eq}.

\begin{equation}\label{four_eq}
   \centering \Phi_{indirect} y_j(x_i) =  \sum_{k=1}^{j-1} \Phi y_j(y_k) \cdot Z_{k}(x_i)
\end{equation}

where $j > 1$ and the function $Z_k(x_i)$ computes the weight vector for all paths from output $y_k$ down to $x_i$. 
For $k > 1$ and $Z_1(x_i)=W(y_1, x_i)$, $Z_k(x_i)$ is recursively computed as follows: 

\begin{equation}\label{six_eq}
    Z_{k}(x_i)=\sum_{l=1}^{k-1} W(y_k, y_{k-l}) \cdot Z_{k-l}(x_i)+W(y_k, x_i)
\end{equation}

where $W(y_k, y_{k-l})$ is the corresponding weight of $y_{k-l}$ to predict the next output $y_k$ (the direct contribution of $y_{k-l}$ to predict $y_k$. 
And, $W(y_k, x_i)$ is the weight of $x_i$ to predict $y_k$ (the direct contribution of $x_i$ to predict $y_k$). The weights $W(y_k, y_{k-l})$ and $W(y_k, x_i)$ are calculated according to:

\begin{equation}\label{sev_eq}
 W(y_{k},.)=\frac{|\Phi y_k(.)|}{\left(\sum_{q=1}^{n}|\Phi y_k(x_q)|+\sum_{p<k}|\Phi y_k(y_p)|\right)}
\end{equation}

where $\Phi y_k(x_q)$ is the direct contribution, as in Eq.~\ref{first_eq}; of each feature $x_q$ to predict $y_k$). $p<k$ means the output $p$ is chained to the output $j$ forming a directed acyclic graph illustrated in Fig.~\ref{fig:CC}. 

For instance, in order to have a complete fair distribution of feature importance for the prediction of $Y_{\mathsf{PSO}}$, we compute the indirect Shapley values of the features  $X_{\mathsf{PA}}$ and $X_{\mathsf{EH}}$. We do so by distributing the direct Shapley value of $Y_{\mathsf{OB}}$ computed previously to the four features. By the distribution operation, we mean the multiplication of the direct Shapley value of each feature by the direct Shapley value of $Y_{\mathsf{OB}}$, divided by the sum of the shapley values of all features for to predict the same output(here $Y_{\mathsf{OB}}$). 

We generalize this mechanism in Algorithm~\ref{alg2} of calculating indirect Shapley values to the chain structure in Fig.~\ref{sv}. The first output $y_1$ has always zero indirect Shapley values because there is no output that precedes it in the chaining. Thus, for the rest of this section, we compute feature indirect contributions for $y_j \in \{y_2, y_3, ..., y_{m}\}$. For each output $y_j$, there exists one direct path to the features thus one direct feature contributions and $2^j-1$ indirect paths for each feature. 
 
\begin{algorithm}
\caption{Computing feature indirect contributions}\label{alg2}
\begin{algorithmic}[1]

\Procedure{inContribution}{$X, Y, \Phi$}  \Comment{inputs, outputs, Shapley values of features and outputs}
    \State $i=j=0$
    \While{$j<len(Y)$} 
        \While{$i<len(X)$}
            \State compute $W(y_k, y_{k-l})$ and $W(y_k, x_i)$ in Eq.~\ref{sev_eq} 
            \State compute $Z_k(x_i)$ in Eq.~\ref{six_eq}
        \EndWhile
    \EndWhile 
    \State return $\Phi_{indirect} y_j(x_i)$ in Eq.~\ref{four_eq} \Comment{returning indirect feature contributions.}
\EndProcedure

\end{algorithmic}
\end{algorithm}

One should notice that for the matter of the simplicity of understanding, we take the absolute value in Eq.~\ref{sev_eq}. Thus, all the contributions will be positive. These absolute values can be replaced by the raw Shapley values in order to keep the positive or negative sign of feature contributions. Keeping the sign helps to understand if the feature penalizes or is in favor of the prediction.

\section{Experiments}
In order to assess the importance of the features that is attributed by our proposed framework\footnote{https://github.com/cwayad/shapleychains} to explain their contributions to predict multiple outputs with a classifier chain, we run experiments on both synthetic and real-world datasets: a $xor$ data that we describe next, and the Adult Income dataset from the UCI repository \cite{Dua:2019}.
Here, we rely on human explanation to validate our results.

\subsection{Synthetic data}\label{synthetic}
To demonstrate our work, we first run experiments on a multi-output synthetic dataset containing two features ($x_1$ and $x_2$) and three outputs ($and$, $or$ and $xor$) corresponding to the logical operations of the same names performed on $x_1$ and $x_2$. 
We split this dataset to 80\% for the training and 20\% for the test of our classifier. 

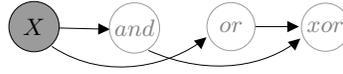
\begin{figure}[!htb]
\centering
\tikzset{every picture/.style={line width=0.3pt}} 
\begin{tikzpicture}[x=0.5pt,y=0.5pt,yscale=-1,xscale=1]
\draw [color={rgb, 255:red, 0; green, 0; blue, 0 }  ,draw opacity=1 ]   (396,138) -- (426,138) ;
\draw [shift={(429,138)}, rotate = 180] [fill={rgb, 255:red, 0; green, 0; blue, 0 }  ,fill opacity=1 ][line width=0.08]  [draw opacity=0] (8.93,-4.29) -- (0,0) -- (8.93,4.29) -- cycle    ;
\draw [color={rgb, 255:red, 0; green, 0; blue, 0 }  ,draw opacity=1 ]   (315,156) .. controls (363.5,177.34) and (405.41,166.69) .. (427.96,148.69) ;
\draw [shift={(430,147)}, rotate = 499.18] [fill={rgb, 255:red, 0; green, 0; blue, 0 }  ,fill opacity=1 ][line width=0.08]  [draw opacity=0] (8.93,-4.29) -- (0,0) -- (8.93,4.29) -- cycle    ;
\draw [color={rgb, 255:red, 0; green, 0; blue, 0 }  ,draw opacity=1 ]   (247,139) -- (283,139.92) ;
\draw [shift={(286,140)}, rotate = 181.47] [fill={rgb, 255:red, 0; green, 0; blue, 0 }  ,fill opacity=1 ][line width=0.08]  [draw opacity=0] (8.93,-4.29) -- (0,0) -- (8.93,4.29) -- cycle    ;
\draw [color={rgb, 255:red, 0; green, 0; blue, 0 }  ,draw opacity=1 ]   (242,153) .. controls (284.68,181.13) and (333.95,165.98) .. (356.94,147.7) ;
\draw [shift={(359,146)}, rotate = 499.18] [fill={rgb, 255:red, 0; green, 0; blue, 0 }  ,fill opacity=1 ][line width=0.08]  [draw opacity=0] (8.93,-4.29) -- (0,0) -- (8.93,4.29) -- cycle    ;
\draw    [fill={rgb, 255:red, 155; green, 155; blue, 155 }  ,fill opacity=1 ] (228.5, 138) circle [x radius= 19.01, y radius= 19.01]   ;
\draw (228.5,138) node    {$\ \ X\ \ $};
\draw  [color={rgb, 255:red, 155; green, 155; blue, 155 }  ,draw opacity=1 ]  (304.5, 139) circle [x radius= 19.01, y radius= 19.01]   ;
\draw (304.5,139) node    {$\textcolor[rgb]{0.5,0.5,0.5}{and}$};
\draw  [color={rgb, 255:red, 155; green, 155; blue, 155 }  ,draw opacity=1 ]  (377.5, 138) circle [x radius= 18.2, y radius= 18.2]   ;
\draw (377.5,138) node    {$\textcolor[rgb]{0.5,0.5,0.5}{\ or\ }$};
\draw  [color={rgb, 255:red, 155; green, 155; blue, 155 }  ,draw opacity=1 ]  (449, 138.44) circle [x radius= 18.2, y radius= 18.2]   ;
\draw (449,138.44) node    {$\textcolor[rgb]{0.5,0.5,0.5}{xor}$};
\label{fig:ccxor}
\end{tikzpicture}
\caption{The classifier chain structure for $xor$ data. $X$ is the set of features $x_1$ and $x_2$. $and$, $or$ and $xor$ are the outputs for which we want to compute direct and indirect Shapley values.} \label{fig:ccxor}
\end{figure}

Next, we construct a classifier chain with the chaining order illustrated in Fig.~\ref{fig:ccxor}. We use a logistic regression as the base learner. Our method is model agnostic meaning that it can be applied to a classifier chain with any other base learners. The use of the logistic regression as the base learner to predict $xor$ is justified by the accuracy that this model achieves compared to other classifiers like decision trees. 
The classifier chain is trained on the train set using $x_1$ and $x_2$ to predict $and$ and $or$ separately. Then, we append these two predicted outputs to the features set in order to predict $xor$. Here, the order in which we predict $and$ and $or$ does not change our method's behavior. 

\begin{figure}[!htb]
    \centering
    \includegraphics[height=4cm,width=5cm]{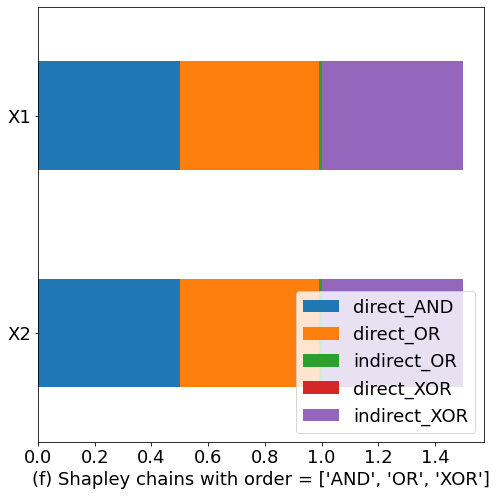}
    \includegraphics[height=4cm,width=5cm]{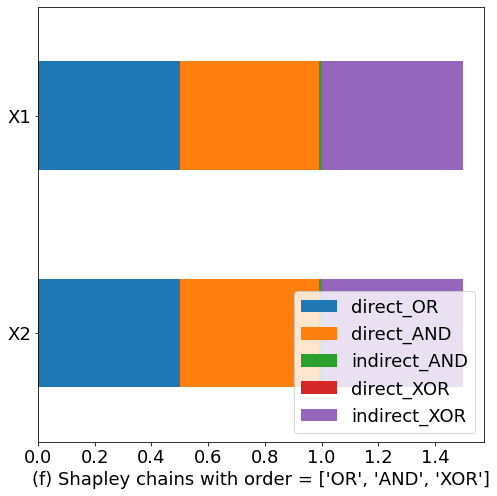}
    \includegraphics[height=4cm,width=5cm]{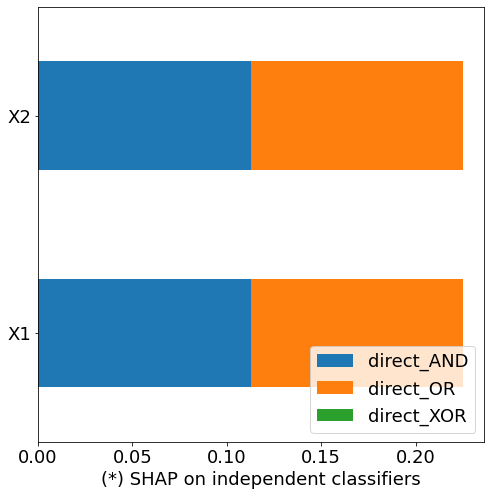}
    \caption{A comparison of SHAP applied on independent classifiers and Shapley Chains. From the left to the right.
    ($a$) and ($b$) Normalized direct and indirect feature contributions made by Shapley Chains to predict $and$, $or$ and $xor$ for chain orders [$and$, $or$, $xor$] and [$or$, $and$, $xor$].
    ($*$) SHAP assigns contributions to $x_1$ and $x_2$ only to predict $and$ and $or$ outputs and completely misses their contributions to predict $xor$.
    Absent colors refer to null Shapley values. }
    \label{fig:xor1}
\end{figure}

To explain the influence of $x_1$ and $x_2$ on the prediction of $xor$, we compared the application of the framework SHAP on each classifier independently and Shapley Chains on the trained classifier chain. We report our analysis on the test data. 
The results of the comparison shown in Fig.~\ref{fig:xor1} indicate that the output chaining propagates the contributions of $x_1$ and $x_2$ to predict $xor$ via $and$ and $or$. 
Specifically, Fig.~\ref{fig:xor1}($a$) and Fig.~\ref{fig:xor1}($b$) illustrate that our method detects the indirect contributions of $x_1$ and $x_2$ (indirect\_xor) to predict $xor$ thanks to the chaining of $and$ and $or$ to $xor$ implemented with the classifier chain model, which tracks down all feature contributions through the chaining of outputs. Furthermore, Fig.~\ref{fig:xor1}($a$) and Fig.~\ref{fig:xor1}($b$) confirm that predicting $or$ before $and$ or vice versa does not affect the feature contributions attribution, which confirms the chain structure for this data.  
On the other hand, these contributions of $x_1$ and $x_2$ are completely neglected by the SHAP framework on independent classifiers (Fig.~\ref{fig:xor1}($*$)).

\subsubsection{Impact of the chaining order on the classifier chain explainability}\label{impact}
\begin{figure}[!htb]
    \centering
    \includegraphics[height=4cm,width=5cm]{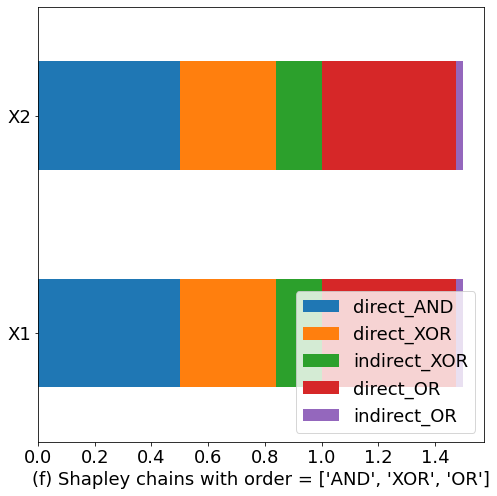}
    \includegraphics[height=4cm,width=5cm]{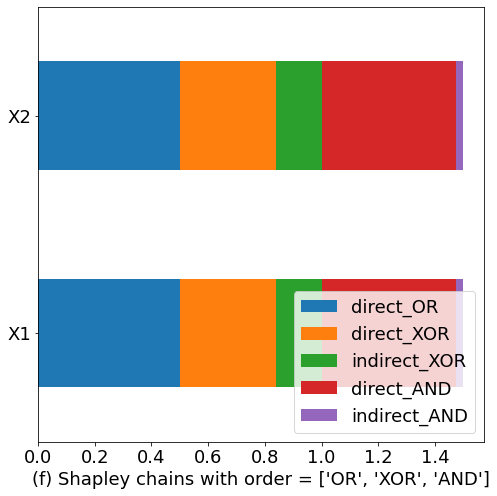}
    \includegraphics[height=4cm,width=5cm]{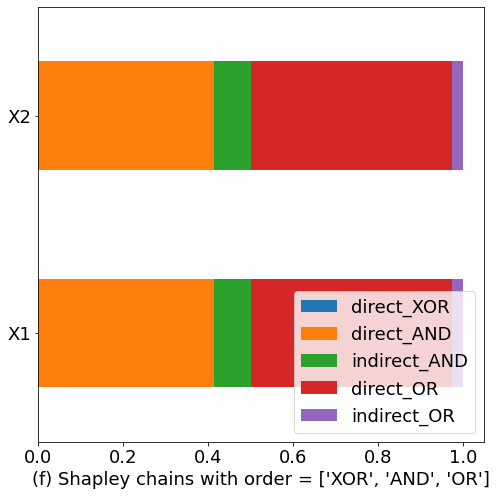}
    \includegraphics[height=4cm,width=5cm]{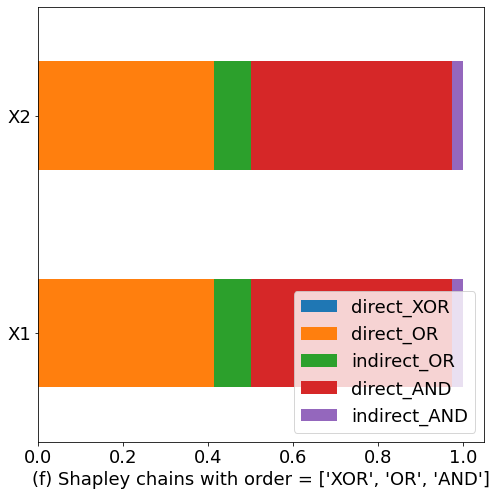}
    \caption{ Possible output chaining orders for $xor$ data. Normalized total feature contributions (direct and indirect Shapley values) for $c$, $d$, $e$ and $f$. }
    \label{fig_comparais}
\end{figure}

In order to measure the impact of the chaining order on the explainability of our classifier chain model with Shapley Chains, we performed analysis on the $3 \,!=6$ possible output chaining orders in the synthetic dataset (scenarios (a) and (b) in Fig.~\ref{fig:xor1} and scenarios (c), (d), (e) and (f) in Fig.~\ref{fig_comparais}).

The information known to the classifier chain when training each output changes depending on the order of these outputs. For instance,
in scenarios $a$ and $b$ (Fig.~\ref{fig:xor1}), we first learn the two outputs $and$ and $or$ using $x_1$ and $x_2$ features. $xor$ is then predicted using $and$ and $or$. 
Here, in both scenarios, both features $x_1$ and $x_2$ contribute indirectly (through $and$ and $or$) to predict $xor$. 
Meanwhile in the scenario $c$ (or $d$), the model relies on $and$(or $or$), $x_1$ and $x_2$ to predict $xor$. We observe that $x_1$ and $x_2$ have direct and indirect contributions, meaning that the classifier chain relies partially on these two features to predict $xor$ (direct contributions of $x_1$ and $x_2$), and on $and$ (indirect contributions of $x_1$ and $x_2$ via $and$). 
The last two scenarios $e$ and $f$ show no contribution of $x_1$ and $x_2$ to predict $xor$, which is explained by the fact that using only these two features, the model can not predict $xor$ without having the information about the dependencies of $xor$ to $and$ and $or$.

These results show that the chain order of $and$, $or$ and $xor$ outputs has an important role in the explainability of the classifier chain, because feeding different inputs to the classifier chain yields different predictions, thus different Shapley values are attributed to the features. 
$x_1$ and $x_2$ importance scores can either be derived from a direct inference of $xor$ output only if there is additional information on output dependencies (for example $and$ is linked to $xor$) or by extracting it from the chain that links $and$ and $or$ to $xor$. 
In the absence of all output dependencies of $and$ or $or$ to $xor$, the model completely ignores the importance of features $x_1$ and $x_2$ in the prediction of $xor$.

\subsection{Explaining Adult Income with Shapley chains}

We run Shapley Chains on the UCI Adult Income dataset. This dataset contains over 32500 instances with 15 features. We first discretize $workclass$, $marital$ $status$ and $relationship$ characteristics. We remove $race$, $education$ and $native$ $country$ and normalize the dataset with the min/max normalizer. Next, we split it into two subsets, using 80\% for the training and the remaining 20\% for testing. We evaluated the hamming loss of a classifier chain with different base learners and we kept the best base classifier, the logistic regression in this case.  

\begin{figure}[!htb]
\begin{minipage}[]{2in}
    \hspace*{\fill}\includegraphics[height=4cm,width=6cm]{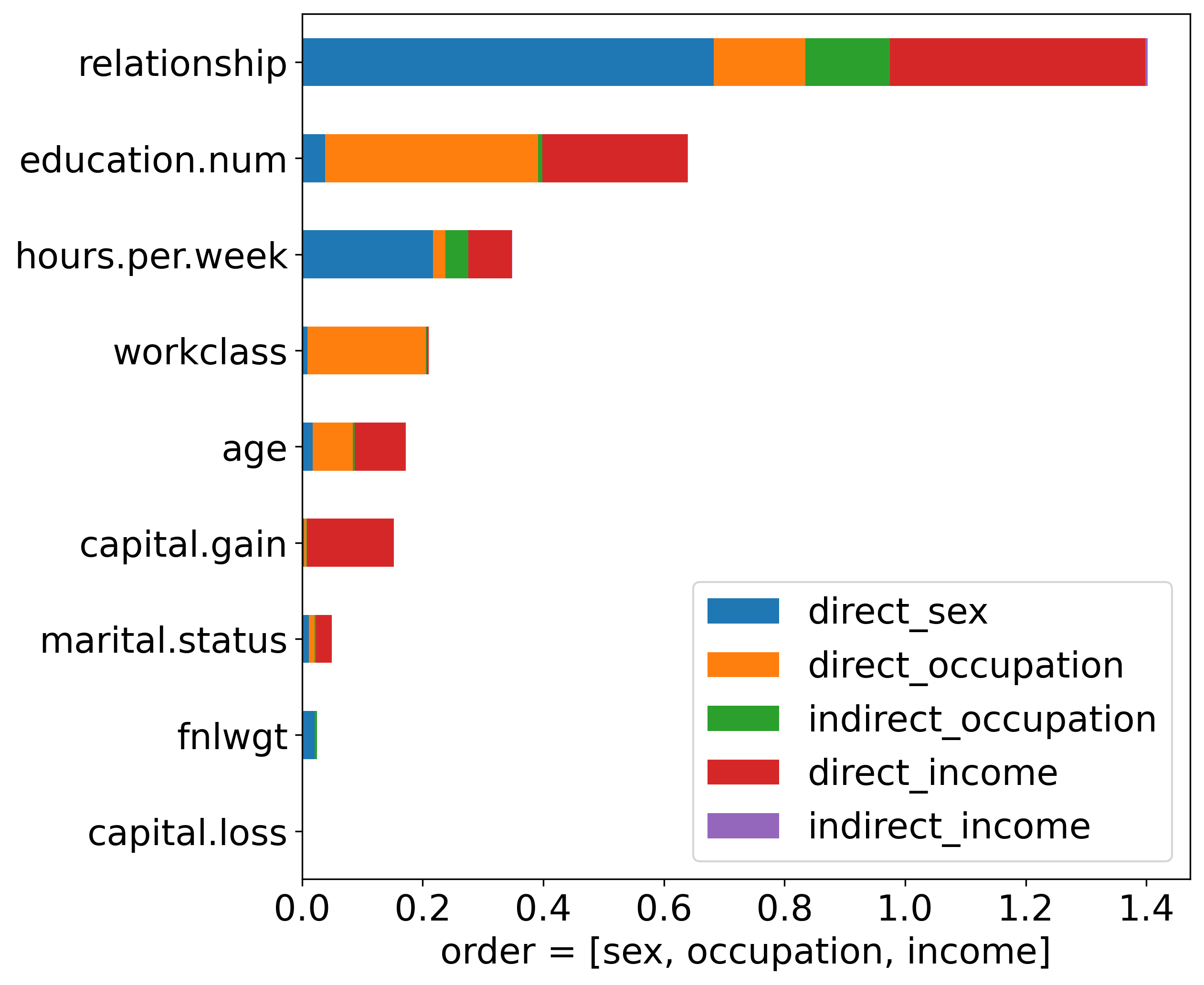}\hspace*{\fill}\par
    \hspace*{\fill} (a) Shapley Chains\hspace*{\fill}
\end{minipage}
\begin{minipage}[!htb]{3in}
    \hspace*{\fill}\includegraphics[height=4cm,width=6cm]{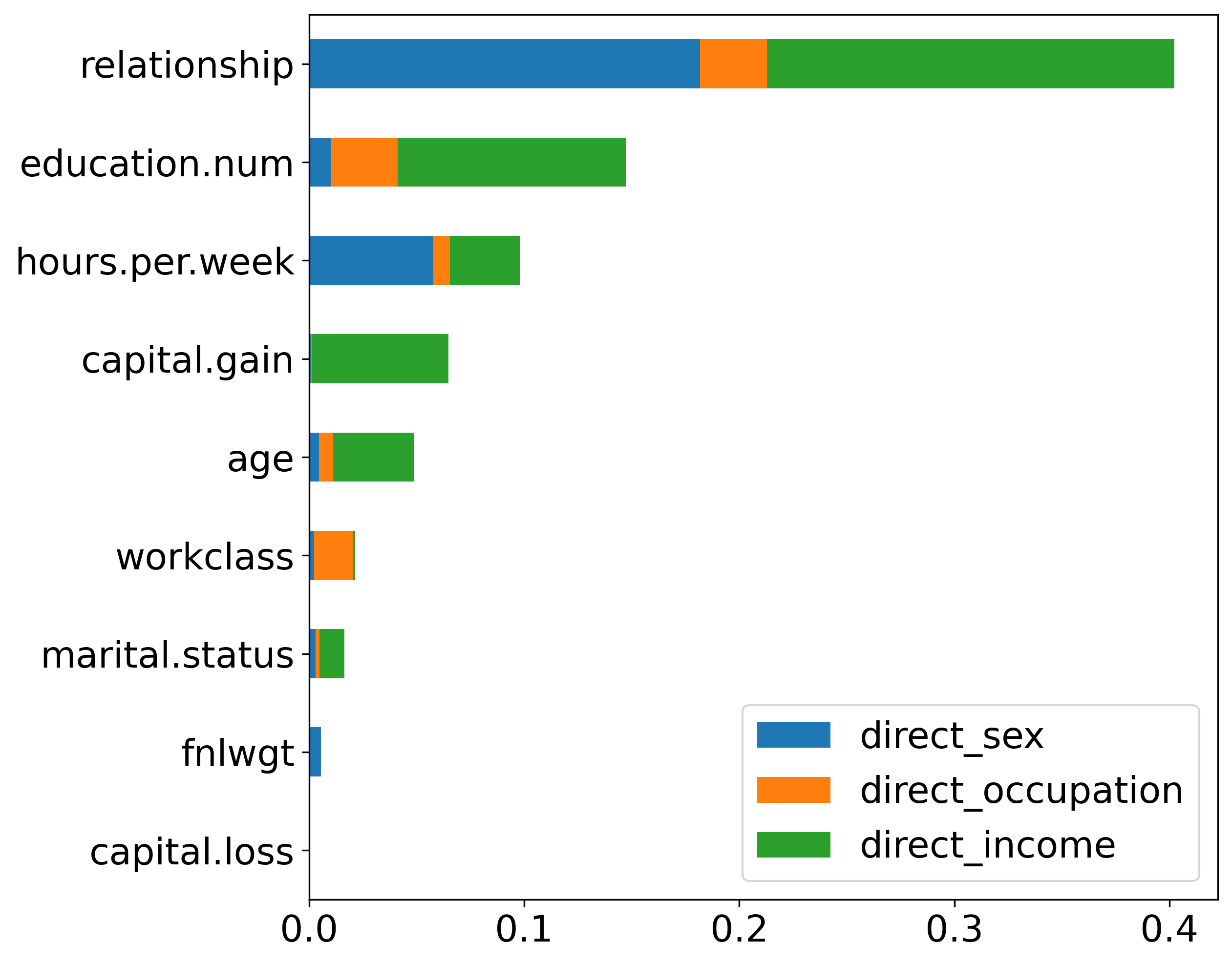}\hspace*{\fill}\par
    \hspace*{\fill} (b) SHAP on independent classifiers\hspace*{\fill}
\end{minipage}
    \caption{(a) Direct and indirect Shapley values on Adult Income data: we normalize and stack each feature's direct and indirect contributions to each output. $sex$ has only direct contributions because it is the first output we predict in this chain order. (b) Stacked Shapley values of independent classifiers on Adult Income data.}
\label{fig:adult}
\end{figure}

In order to explain feature contributions to the predictions of the three outputs $sex$, $occupation$ and $income$, we compared the results of Shapley Chains against classic Shapley values applied on separate logistic regression classifiers for different chain orders. 
Fig.~\ref{fig:adult} shows graphical representation of normalized and stacked feature contributions when applying Shapley Chains on our data set (Fig.~\ref{fig:adult}.(a)), and stacked feature contributions from independent logistic regression classifiers (Fig.~\ref{fig:adult}.(b)).
In both cases, the magnitude of the feature contributions is greater in Shapley Chains compared to independent Shapley values, which confirms our initial hypothesis of some contributions are missed by SHAP framework, and these contributions can be detected when we take into account output dependencies. 
For example, the number of hours worked in a week ($hours.per.week$) has a more important indirect contribution to predict individual's $occupation$ than a direct contribution. This is explained by the fact that $sex$ is related to $occupation$, and this relationship is propagated to the features by Shapley Chains. $relationship$ is another example of Shapley Chains detecting indirect feature contributions to predict $occupation$.
Furthermore, feature rankings are different in Shapley Chains. For example, the ranking of $capital.gain$ comes in the fourth position (before $workclass$) using SHAP applied to independent classifiers. In our method, this feature's ranking is always less important (according to different chaining orders) than $workclass$ to predict $sex$, $occupation$ and $income$ which makes more sens to us. 

\begin{figure}[!htb]
    \centering
    \begin{minipage}[]{1.5in}
    \includegraphics[height=4cm,width=4cm]{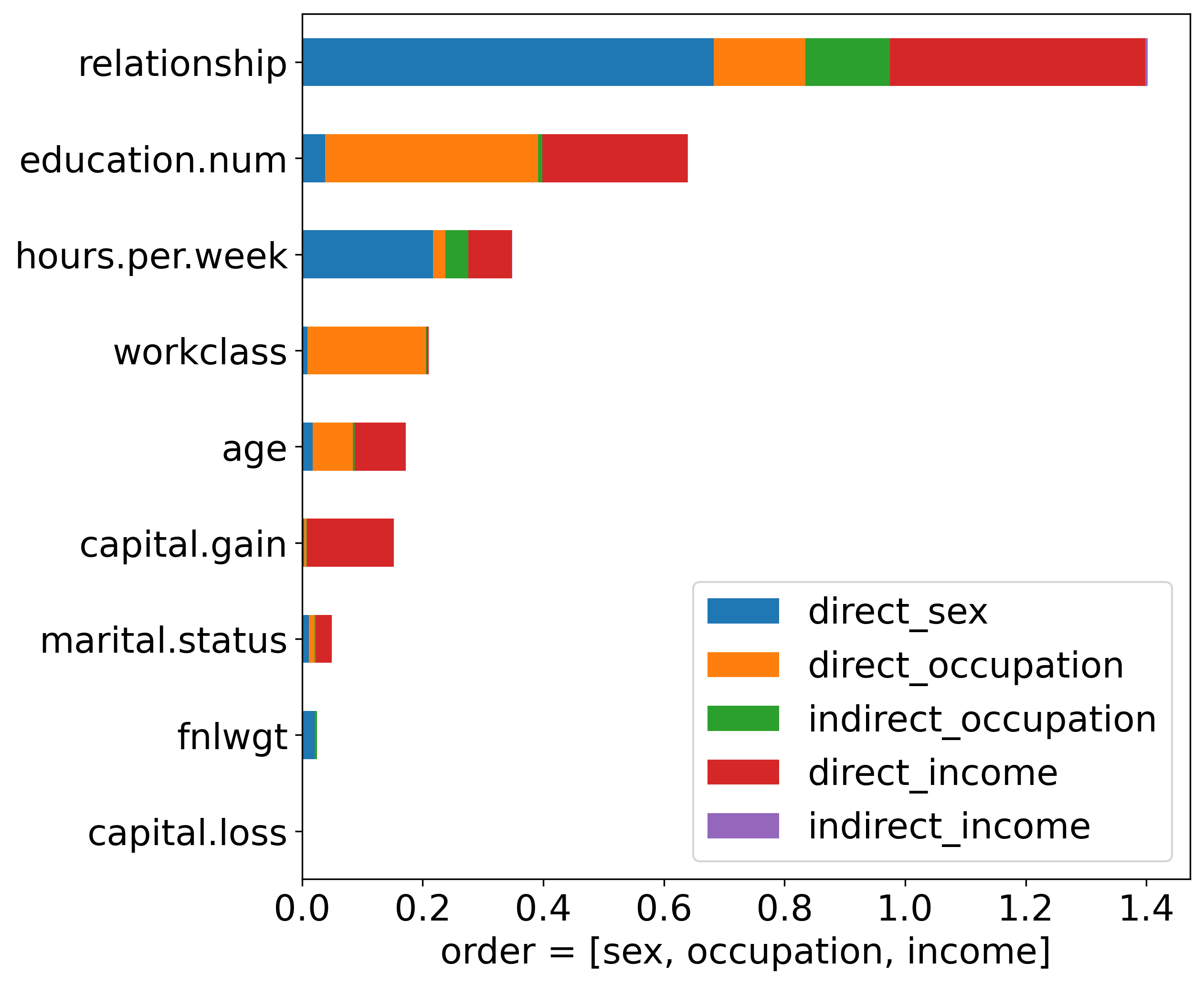}
    \hspace*{\fill}\par
    \hspace*{\fill} (a)\hspace*{\fill}
    \end{minipage}
    \begin{minipage}[]{1.5in}
    \includegraphics[height=4cm,width=4cm]{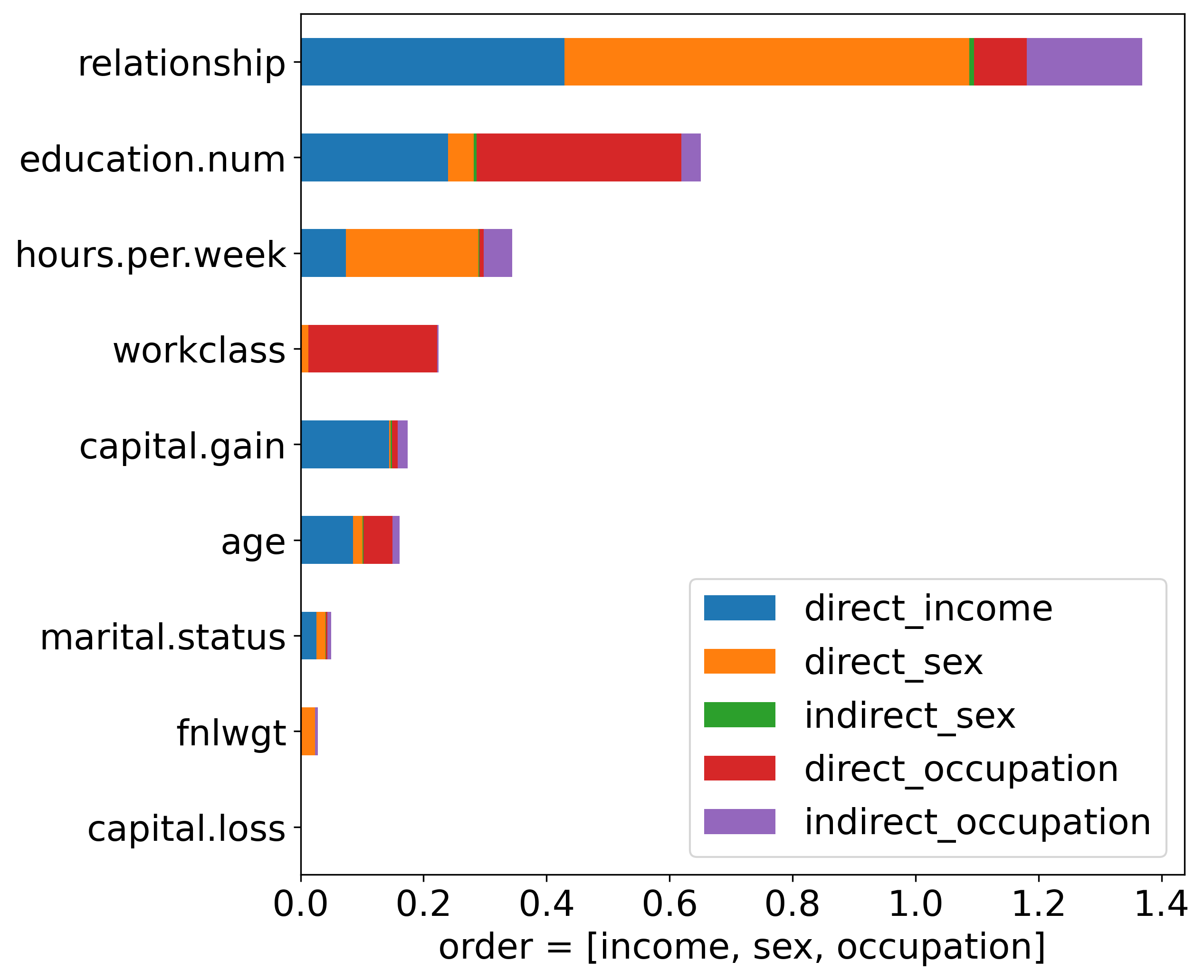}
    \hspace*{\fill}\par
    \hspace*{\fill} (b)\hspace*{\fill}
    \end{minipage}
    \begin{minipage}[]{1.5in}
    \includegraphics[height=4cm,width=4cm]{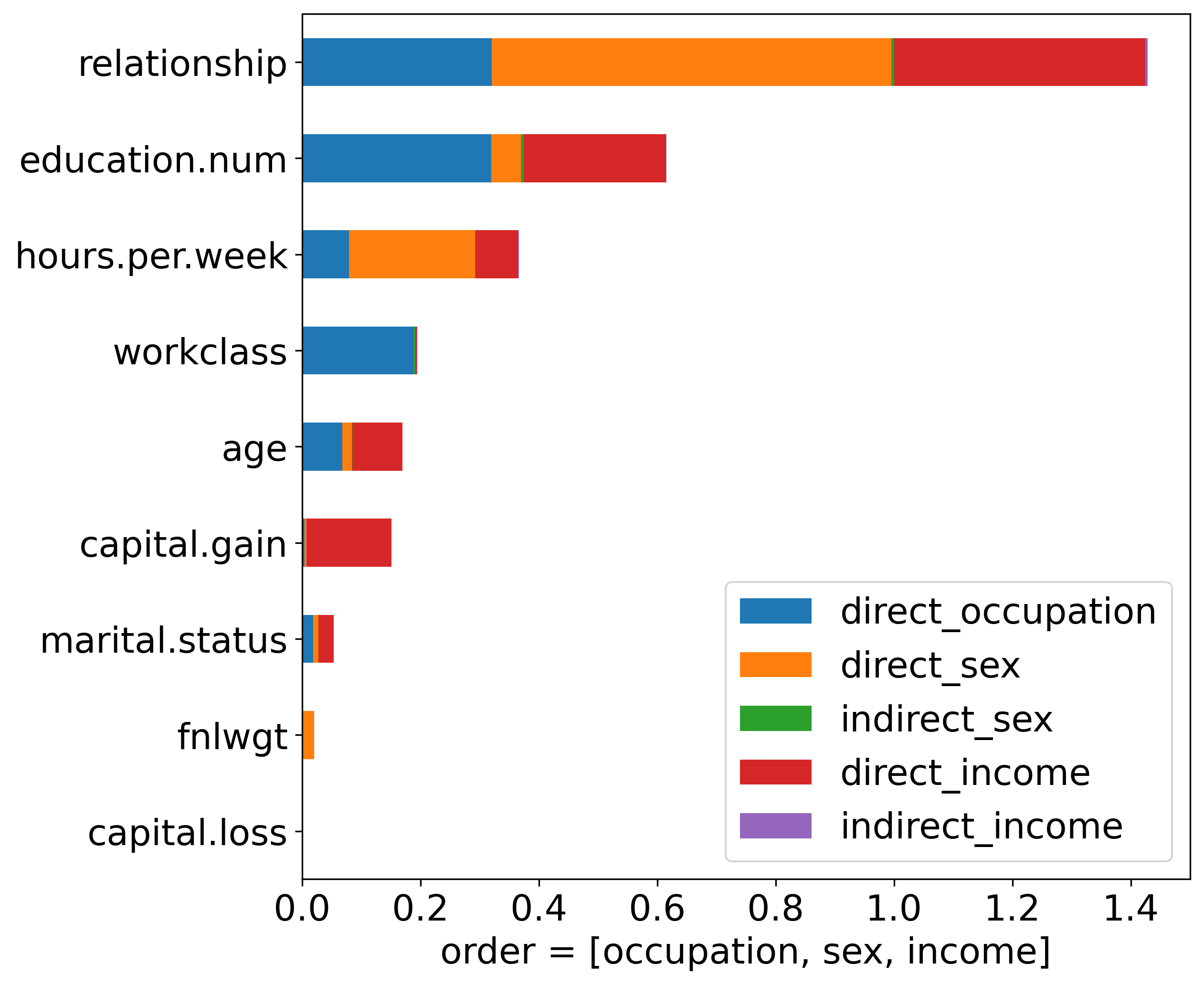}
    \hspace*{\fill}\par
    \hspace*{\fill} (c)\hspace*{\fill}
    \end{minipage}
	\caption{Stacked direct and indirect feature effects for 3 different chain structures over Adult Income data.}
    \label{fig:adult2}
\end{figure}

We also tested the impact of different chain orders of these three outputs on the feature importance attribution. Fig.~\ref{fig:adult2} illustrates three different chaining orders. Each different order allows each classifier to use different prior knowledge to learn these outputs. For example in Fig.~\ref{fig:adult2}(b), we first predict $income$ and $sex$ and we use this information to predict $occupation$. 
Intuitively, $occupation$ is correlated to individual’s $sex$ and $income$.
The classifier chain uses this information provided to the third classifier to predict $occupation$. Here, Shapley Chains attribute more importance to the factors that predict both $income$ and $sex$, when predicting $occupation$. 
Shapley Chains preserve the order of feature importance scores across all the chaining orders in general, but the magnitude of each feature's importance differs from one chain to another. This is due to the prior knowledge that is fed into the classifier when learning each output. In addition, these feature importance scores are always more important in Shapley Chains compared to Shapley values of independent classifiers for all chain orders.  

\section{Conclusions and Perspectives}

In this paper, we presented Shapley Chains, a novel method for calculating feature importance scores based on Shapley values for multi-output classification with a classifier chain. We defined direct and indirect contribution and demonstrated on synthetic and real-world data how the attribution of indirect feature contribution to the prediction is more complete with Shapley Chains.
Our method helps practitioners to better understand hidden influence of the features on the outputs by detecting indirect feature contributions hidden in output dependencies. Although the rankings of feature importance are not always different from independent feature importance scores, the magnitude of these scores is always important in Shapley Chains, which is more important to look at in applications that are sensitive to the magnitude of these importance scores rather than their rankings.
By extending the Shapley value to feature importance attribution of classifier chains, we make use of output interdependencies that is implemented in classifier chains in order to represent the real learning factors of a multi-output classification task. 

To extend this work, Shapley Chains could be evaluated on multi-output regression tasks. Exploring the relationship's type between the outputs, and studying wether Shapley Chains preserve all these relationships when attributing feature contributions is another open question of our work.

\newpage
\bibliographystyle{splncs04}

\begin{thebibliography}{10}
\providecommand{\url}[1]{\texttt{#1}}
\providecommand{\urlprefix}{URL }
\providecommand{\doi}[1]{https://doi.org/#1}

\bibitem{Dua:2019}
Dua, D., Graff, C.: {UCI} machine learning repository (2017),
  \url{http://archive.ics.uci.edu/ml}

\bibitem{fryeShapleyExplainabilityData2021}
Frye, C., {de Mijolla}, D., Begley, T., Cowton, L., Stanley, M., Feige, I.:
  Shapley explainability on the data manifold (Dec 2021)

\bibitem{fryeAsymmetricShapleyValues2021}
Frye, C., Rowat, C., Feige, I.: Asymmetric {{Shapley}} values: Incorporating
  causal knowledge into model-agnostic explainability (Dec 2021)

\bibitem{jensenPsoriasisObesity2016}
Jensen, P., Skov, L.: Psoriasis and {{Obesity}}. Dermatology (Basel,
  Switzerland)  \textbf{232}(6),  633--639 (2016)

\bibitem{lundbergUnifiedApproachInterpreting2017a}
Lundberg, S., Lee, S.I.: A {{Unified Approach}} to {{Interpreting Model
  Predictions}} (Nov 2017)

\bibitem{montavonLayerWiseRelevancePropagation2019}
Montavon, G., Binder, A., Lapuschkin, S., Samek, W., M{\"u}ller, K.R.:
  Layer-{{Wise Relevance Propagation}}: {{An Overview}}. In: Samek, W.,
  Montavon, G., Vedaldi, A., Hansen, L.K., M{\"u}ller, K.R. (eds.) Explainable
  {{AI}}: {{Interpreting}}, {{Explaining}} and {{Visualizing Deep Learning}},
  vol. 11700, pp. 193--209. {Springer International Publishing}, {Cham} (2019)

\bibitem{readClassifierChainsMultilabel2011}
Read, J., Pfahringer, B., Holmes, G., Frank, E.: Classifier chains for
  multi-label classification. Machine Learning  \textbf{85}(3),  333--359 (Dec
  2011)

\bibitem{readClassifierChainsReview2021}
Read, J., Pfahringer, B., Holmes, G., Frank, E.: Classifier {{Chains}}: {{A
  Review}} and {{Perspectives}}. Journal of Artificial Intelligence Research
  \textbf{70},  683--718 (Feb 2021)

\bibitem{ribeiroWhyShouldTrust2016a}
Ribeiro, M.T., Singh, S., Guestrin, C.: "{{Why Should I Trust You}}?":
  {{Explaining}} the {{Predictions}} of {{Any Classifier}} (Aug 2016)

\bibitem{rozemberczkiShapleyValueClassifiers2021a}
Rozemberczki, B., Sarkar, R.: The {{Shapley Value}} of {{Classifiers}} in
  {{Ensemble Games}} (Jun 2021)

\bibitem{shrikumarLearningImportantFeatures2019a}
Shrikumar, A., Greenside, P., Kundaje, A.: Learning {{Important Features
  Through Propagating Activation Differences}} (Oct 2019)

\bibitem{sundararajanAxiomaticAttributionDeep2017a}
Sundararajan, M., Taly, A., Yan, Q.: Axiomatic {{Attribution}} for {{Deep
  Networks}} (Jun 2017)

\bibitem{wangShapleyFlowGraphbased2021}
Wang, J., Wiens, J., Lundberg, S.: Shapley {{Flow}}: {{A Graph-based Approach}}
  to {{Interpreting Model Predictions}}. In: Proceedings of {{The}} 24th
  {{International Conference}} on {{Artificial Intelligence}} and
  {{Statistics}}. pp. 721--729. {PMLR} (Mar 2021)

\end{thebibliography}

\end{document}